\documentclass[10pt,a4paper]{article}
\usepackage{float}
\usepackage[all]{xy}
\usepackage[affil-it]{authblk}

\usepackage{amsthm}
\usepackage{authblk}

\theoremstyle{definition}

\usepackage[table,xcdraw]{xcolor}
\usepackage{bbm}
\usepackage{hyperref}
\usepackage[utf8]{inputenc}
\usepackage{amsmath}
\usepackage{amsfonts}
\usepackage{amssymb}
\usepackage{enumerate}
\usepackage{amsthm}
\title{A signature-based machine learning model for bipolar disorder and borderline personality disorder}
\usepackage{graphicx}
\author[1]{Imanol Perez Arribas\thanks{imanol.perez@maths.ox.ac.uk}}
\author[2,3]{Kate Saunders\thanks{kate.saunders@psych.ox.ac.uk}}
\author[2,3]{Guy Goodwin\thanks{guy.goodwin@psych.ox.ac.uk}}
\author[1,4]{Terry Lyons\thanks{tlyons@maths.ox.ac.uk}}

\affil[1]{Mathematical Institute, University of Oxford}
\affil[2]{Department of Psychiatry, University of Oxford}
\affil[3]{Oxford Health NHS Foundation Trust}
\affil[4]{Alan Turing Institute}

\usepackage{indentfirst}

\usepackage{afterpage}

\newcommand\restr[2]{{
  \left.\kern-\nulldelimiterspace 
  #1 
  \vphantom{\big|} 
  \right|_{#2} 
  }}

\newcount\colveccount
\newcommand*\colvec[1]{
        \global\colveccount#1
        \begin{pmatrix}
        \colvecnext
}
\def\colvecnext#1{
        #1
        \global\advance\colveccount-1
        \ifnum\colveccount>0
                \\
                \expandafter\colvecnext
        \else
                \end{pmatrix}
        \fi
}

\usepackage[table,xcdraw]{xcolor}
\usepackage{subcaption}

\usepackage{slashbox}

\begin{document}
\maketitle

\begin{abstract}
\textbf{Background:} Mobile technologies offer opportunities for higher resolution monitoring of health conditions. This opportunity seems of particular promise in psychiatry where diagnoses often rely on retrospective and subjective recall of mood states. However, getting actionable information from these rather complex time series is challenging, and at present the implications for clinical care are largely hypothetical. This research demonstrates that, with well chosen cohorts (of bipolar disorder, borderline personality disorder, and control) and modern methods, it is possible to objectively learn to identify distinctive behaviour over short periods (20 reports) that effectively separate the cohorts. Performing this analysis with an individual removed, individuals could then be reevaluated on a spectrum according to how consistent their behaviour was with the different 20 report behaviour patterns of the cohorts.  The spectral analysis provided consistency with the original allocation to cohorts.  This suggests the original clinical selection was successful at identifying different behaviour models and that the methods expanded below could be the basis for an ongoing monitoring of individuals on this spectrum.

\textbf{Methods:} Participants with bipolar disorder or borderline personality disorder and healthy volunteers completed daily mood ratings using a bespoke smartphone app for up to a year. A signature-based machine learning model was used to classify participants on the basis of the interrelationship between the different mood items assessed and to predict subsequent mood.
 
\textbf{Results:} The signature methodology was significantly superior to earlier statistical approaches applied to this data in distinguishing the participant three groups, clearly placing 75\% into their original groups on the basis of their reports. Subsequent mood ratings were correctly predicted with greater than 70\% accuracy in all groups. Prediction of mood was most accurate in healthy volunteers (89-98\%) compared to bipolar disorder (82-90\%) and borderline personality disorder (70-78\%).

\textbf{Conclusion:} The signature approach is an effective approach to the analysis of mood data, which, despite the scale of the data, was able to capture useful information. The three cohorts offered internally consistent but distinct patterns of mood interaction in their reporting; the original selection identified objectively different communities whose behaviour patterns might act as a useful comparator for future monitoring.  

\end{abstract}

\section{Introduction}\label{sec:introduction}

The rapid expansion in mobile technologies over the last ten years has provided a platform for much more accurate phenotyping of psychiatric disorders. Historically diagnosis has relied on an anamnestic approach and has been hampered by the inherent inaccuracy of retrospective recall of mood states. Momentary assessment using mobile phones and other wearables has revealed more precise measures of psychopathology and highlighted the shortcomings of current diagnostic categories. The NIMH Research Domain Criteria (RDoC) propose a new data-driven approach \textit{bottom up} approach to diagnosis. The oscillatory nature of psychiatric symptomatology poses a significant analytic challenge. Rough path theory and the signature method provide a means of analysing complex ordered data and specifically streams of data that are highly oscillatory in nature. In \cite{GUY} for instance, the authors study a method to transform a stream of data into an associated rough path. The signature of a continuous path is an infinite sequence that summarises somehow the information  contained in the path (\cite{LYOSPRINGER}). Signatures have many properties which can be used as features in standard machine learning approaches (see \cite{FUTURE} and \cite{MACHINELEARNING}).

In this paper, we use a signature-based machine learning model to analyse data obtained from a clinical study in \cite{PSYCHIATRY}, which explored daily reporting of mood in participants with bipolar disorder, borderline personality disorder and healthy volunteers. Specifically we used a signature-based method to classify participants on the basis of their mood and then to predict their mood the following day. Although these questions are usually seen as two different problems from a supervised learning perspective – the first question is a classification problem, and the second a time series forecasting problem – the generality of the signature-based machine learning model allows these problems to be treated in similar ways.

\section{Methods}

\subsection{Data}\label{sec:data}

Prospective mood data was captured from 130 individuals who were taking part in the AMoSS study. 48 participants in the cohort were diagnosed with bipolar disorder, 31 were diagnosed with borderline personality and 51 participants were healthy. Self-reported mood scores were recorded daily using a bespoke smartphone app. Each day participants rate their mood across six different categories (anxiety, elation, sadness, anger, irritability and energy) using a 7-point Likert scale (with values from 1 (not at all) to 7 (very much)). Data was collected from each participant for a minimum of 2 months, although 61 of the 130 participants provided data for more than 12 months. Overall compliance was 81.2\% overall.

The mood data was divided into streams of 20 consecutive observations. Although these observations are typically daily observations, since some participants failed to record their mood in some days the stream of data doesn't need to come from 20 consecutive days. This generated 733 streams of data. These were randomly separated into a training set of 513 streams of data, and a testing set of 220 streams of data. We then normalised the data in order to apply a signature-based machine learning model. Normalisation allows to make an abstraction of the data that ignores some of the redundant information.

\begin{figure}[H]
\centering
\includegraphics[width=\linewidth]{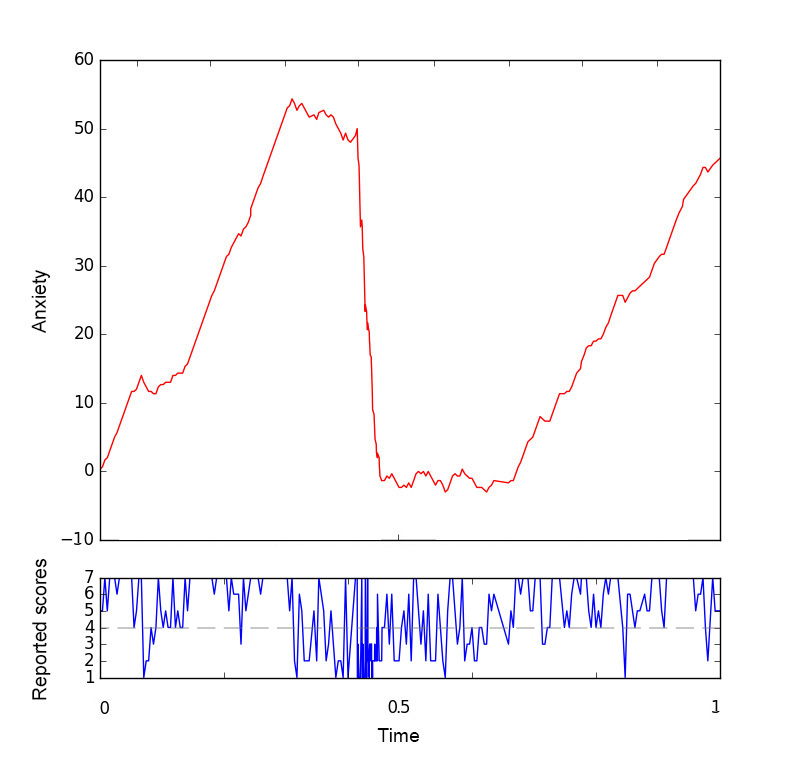}

\caption{Normalised anxiety scores of a participant with bipolar disorder (above), which were calculated using the reported scores (below). As we see, high levels of reported scores correspond to upward trends, low levels of reported scores correspond to downward trends and periods of time of high oscillations in the reported scores are represented by oscillations in the path.}
\label{fig:normalised scores}
\end{figure}

The streams of mood reports were identified as a 7-dimensional path \linebreak$\{(t_i,S_{t_i})\}^{19}_{ i=0}$ (1 dimension for time, 6 dimensions for the scores) defined on the unit interval. Figure \ref{fig:normalised scores} shows the normalised anxiety scores of a participant with bipolar disorder. Intuitively, if the path that consists of time against the scores of one of the categories has an upward trend, the participant has scores greater than 4 in that category. If the trend is downward, on the other hand, the participant had a score lower than 4 in that category. Hence, if the mood of a particular participant changes drastically from day to day, the normalised path should also oscillate considerably. If the mood is very stable, however, the corresponding path will have stable trends.

\begin{figure}[H]
\centering
\includegraphics[width=\linewidth]{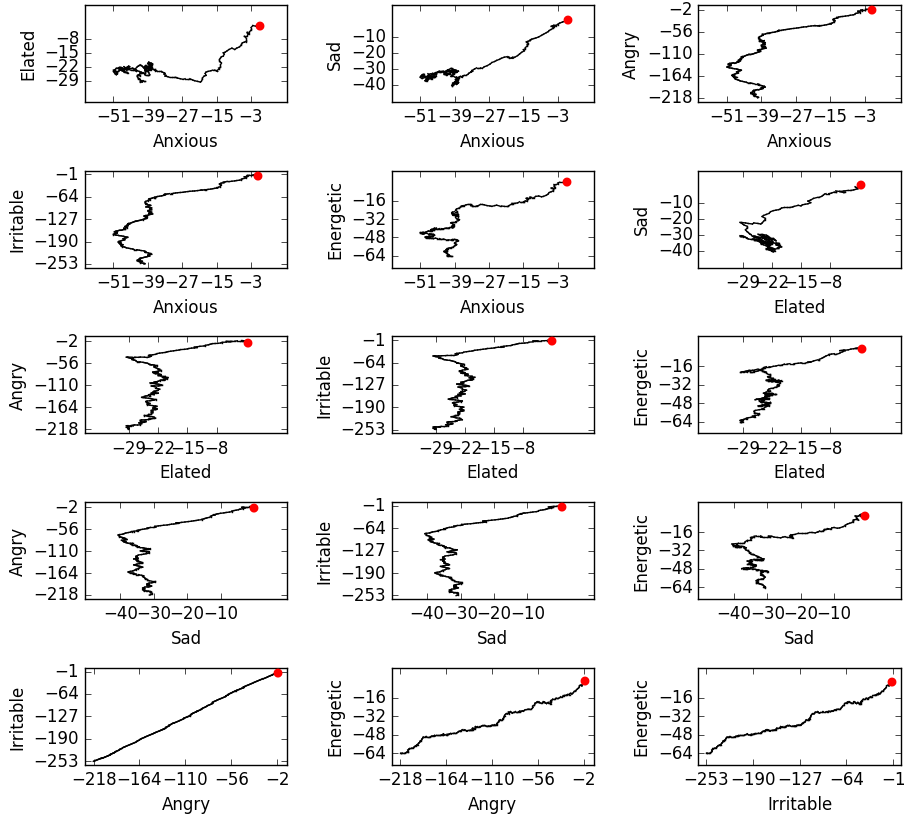}

\caption{Normalised scores of each category plotted against all other categories, for a participant with bipolar disorder. The red point indicates the starting point. Notice that the scale is different in each plot.}
\label{fig:all against all}
\end{figure}

Figure \ref{fig:all against all}, on the other hand, shows the normalised scores of each category plotted against all other categories. This way of representing the path allows to interpret the order at which the participant changes of mood. Take for example, the Angry vs Elated plot (third row, first column). The path starts at the point (0, 0). As we see, the path moves left first. Therefore, the participant is becoming less and less elated, while the score in anger remains approximately constant. Suddenly, the period of low elation stops, and the participant starts recording low scores in anger. These low levels in anger remain persistent for the rest of the path. On the other hand, if we take a look to the Angry vs Irritable plot (fifth row, first column), we observe that it is essentially a straight line. Therefore, the levels of anger and irritability are highly correlated and the scores in both categories are roughly the same, for this participant.

\subsection{Group classification}\label{subsec:group classification}

With the objective of building a machine learning model that classifies each participant with the corresponding clinical group, we started establishing a set of input-output pairs $\{(R_i,Y_i)\}_i$, where $R_i$ is the 7-dimensional path of participant $i$, as described above, and $Y_i$ denotes the group it was diagnosed into at the beginning of the study. This set was transformed into a new set of input-output pairs, $\{(S^n(R_i),Y_i)\}_i$, where $S^n(R_i)$ denotes the truncated signature of order $n$ of the stream $R_i$ (see \cite{FUTURE}). The size of the truncated signature of order $n$ grows exponentially with $n$, and therefore large values of $n$ will produce input vectors of large dimensions. However, in practice we will only considered $n = 2,3,4$.

Given that it has already been established that there are differences in mean mood scores between the groups overall (\cite{PSYCHIATRY}) we also calculated the mean score in each mood category and classified streams of 20 consecutive observations on this basis as a comparison to the signature method. To assess the performance of the classification procedure we computed accuracy, sensitivity, specificity, positive predicted value (PPV). We used the area under the receiver operating characteristic curve (RoC) to assess the performance of the classification model at different values of threshold. Moreover, in order to have a better understanding of how robust these percentages are, we applied bootstrapping to the model with the order fixed to 2.

We also considered the extent to which participants where characterised as belonging to each specific clinical group. We trained the model using all participants but the one we are interested in, and then test the model with 20-observations periods from this person. We then calculated the proportion of periods of time when the participant was classified as bipolar, healthy or borderline, which allows us to plot the participant as a point in the triangle. This process was followed for every participant.

\subsection{Predicting the mood of a participant}\label{subsec:prediction}

In this case, the aim is to predict the mood of a participant, using the last 20 observations of the participant. We constructed the input-output pairs $\{(R_i,Y_i)\}_i$, where $R_i$ is the normalised 7-dimensional path of a particular participant, and $Y_i \in \{1,\ldots,7\}^6$ is the mood of the participant the next observation he or she will register. We then constructed a new set of input-output pairs $\{(S^n(R_i),Y_i)\}_i$, with $S^n(R_i$) the truncated signature of order $n\in \mathbb{N}$. We applied regression using these pairs of inputs and outputs, obtaining thus our model. The model was trained using data from each clinical group separately.

In order to measure the accuracy of the predictions, we used the Mean Absolute Error (MAE) and the percentage of correct predictions. Since we were interested in predicting the mood of the participant (that is, how anxious, energetic, sad, etc. he or she will be in the future) we are not interested in correctly predicting the exact score in a particular category. After all, if the predicted score is close to the observed score then the mood was correctly predicted, even if the observed and predicted scores do not coincide. Therefore, a prediction $\widehat{y}\in \{1, \ldots, 6\}$ will be considered correct if $|y-\widehat{y}|\leq 1$, where $y$ is the correct score, since in this case the prediction correctly captured the mood of the participant.

We used the publicly available software eSig to calculate signatures of streams of data, Python pandas package (version 0.20.1) for statistical analysis, data manipulations and processing, Python scikit-learn package (version 0.18.1) for machine learning tasks and matplotlib for plotting and graphics (version 2.0.1).

The study was approved by the NRES Committee East of England – Norfolk (13/EE/0288) and the Research and Development department of Oxford Health NHS Foundation Trust.

\section{Results}\label{sec:results}

\subsection{Predicting which clinical group a participant belongs to}\label{subsec:classify}

\begin{figure}[h]
\centering
\includegraphics[width=\linewidth]{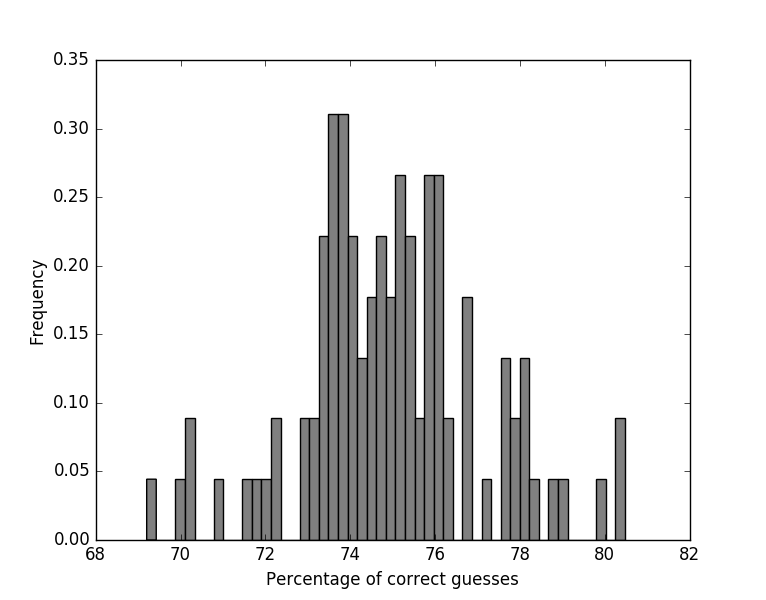}

\caption{Histogram of the percentage of correct predictions obtained after applying bootstrapping. We created a new training set sampling the original training set with replacement, in order to obtain a set of the same size. This process was repeated 100 times, and for each iteration the percentage of correct predictions was recorded in order to obtain the histogram then.}
\label{fig:histogram}
\end{figure}

The signature-based model described in Section \ref{subsec:group classification} categorised 75\% of participants into the correct diagnostic group (Table \ref{table:correct guesses classifier}). The signature method performed significantly better than the naive model using the average in each category over the 20 observations, which classified just 54\% of participants correctly. On bootstrapping the average was 74.85\% while the standard deviation was 2.05, as shown in Figure \ref{fig:histogram}, suggesting that the results are stable and robust.

\begin{table}[]
\centering
\begin{tabular}{cc}
Order & Correct predictions \\ \hline
2nd   & 75\%            \\
3rd   & 70\%            \\
4th   & 69\%           
\end{tabular}
\caption{Percentage of people correctly classified in the three clinical groups, for different orders of the model. The model performed best when the \textit{order} of the model was fixed to 2.}
\label{table:correct guesses classifier}
\end{table}

\begin{table}[]
\centering
\begin{tabular}{
>{\columncolor[HTML]{EFEFEF}}l |lll|l}
\backslashbox{Model prediction}{\hspace{-1cm}Actual clinical group}
       & \multicolumn{1}{l|}{\cellcolor[HTML]{EFEFEF}Borderline} & \multicolumn{1}{l|}{\cellcolor[HTML]{EFEFEF}Healthy} & \cellcolor[HTML]{EFEFEF}Bipolar & \cellcolor[HTML]{EFEFEF}Total \\ \hline
Borderline & \cellcolor[HTML]{9AFF99}37                          & \cellcolor[HTML]{FFCCC9}1                        & \cellcolor[HTML]{FFCCC9}14   & 52                        \\ \cline{1-1}
Healthy    & \cellcolor[HTML]{FFCCC9}4                          & \cellcolor[HTML]{9AFF99}68                       & \cellcolor[HTML]{FFCCC9}9   & 81                        \\ \cline{1-1}
Bipolar    & \cellcolor[HTML]{FFCCC9}14                             & \cellcolor[HTML]{FFCCC9}14                        & \cellcolor[HTML]{9AFF99}59    & 87                        \\ \hline
Total      & 55                                                  & 83                                               & 82                          & 220
\end{tabular}
\caption{Confusion matrix of the predictions. Green cells indicate incorrect predictions, while blue cells show correct predictions.}
\label{table:summary classifier}

\end{table}

\begin{table}[]
\centering
\label{my-label}
\begin{tabular}{|l|l|l|l|}
\hline
           & Healthy                                         & Bipolar                                         & Borderline                                      \\ \hline
Healthy    & \cellcolor[HTML]{333333}{\color[HTML]{343434} } & 84\%                                            & 93\%                                            \\ \hline
Bipolar    & \cellcolor[HTML]{333333}{\color[HTML]{343434} } & \cellcolor[HTML]{333333}{\color[HTML]{343434} } & 80\%                                            \\ \hline
Borderline & \cellcolor[HTML]{333333}{\color[HTML]{343434} } & \cellcolor[HTML]{333333}{\color[HTML]{343434} } & \cellcolor[HTML]{333333}{\color[HTML]{343434} } \\ \hline
\end{tabular}
\caption{Percentage of correct predictions.}
\label{table: pairs percentages}
\end{table}

\begin{table}[]
\centering
\label{my-label}
\begin{tabular}{|l|l|l|l|}
\hline
           & Healthy                                         & Bipolar                                         & Borderline                                      \\ \hline
Healthy    & \cellcolor[HTML]{333333}{\color[HTML]{343434} } & 0.91                                            & 0.98                                            \\ \hline
Bipolar    & \cellcolor[HTML]{333333}{\color[HTML]{343434} } & \cellcolor[HTML]{333333}{\color[HTML]{343434} } & 0.86                                            \\ \hline
Borderline & \cellcolor[HTML]{333333}{\color[HTML]{343434} } & \cellcolor[HTML]{333333}{\color[HTML]{343434} } & \cellcolor[HTML]{333333}{\color[HTML]{343434} } \\ \hline
\end{tabular}
\caption{Area under ROC curve.}
\label{table:pairs roc}
\end{table}

Incorrect predictions differed between the three groups (see Table \ref{table:summary classifier}). Incorrect predictions were low in healthy participants in comparison to participants with bipolar disorder or borderline personality disorder. The model very clearly distinguishes healthy participants from the clinical groups (Table \ref{table: pairs percentages} and \ref{table:pairs roc}).

The proportion of time in which participant's mood appeared characteristic of each specific group -- as described in Section \ref{subsec:group classification} -- is shown in Figures \ref{fig:triangle1}, \ref{fig:triangle2} and \ref{fig:triangle3}. In each of the plots, the regions of highest density of participants are located in the correct corner of the triangle but this is much more consistent in the healthy participants than the two clinical groups.

\subsection{Predicting the mood of a participant}

The model described in Section \ref{subsec:prediction} correctly predicted the next mood score in healthy participants with 89-98\% accuracy. In bipolar participants the mood was correctly predicted 82-90\% of the time while in borderline participants this was 70-78\% of the time (see Table \ref{table:predict mood}).

We compared our model to the obvious benchmark that predicts that next day's score will be equal to the last day's score. In this case, the mood of healthy participants was correctly predicted 61-92\% of the time, while the mood of bipolar and borderline participants was correctly predicted 46-67\% and 44-62\% of the time. Therefore, our model clearly outperformed this benchmark.

\begin{figure}[H]
\centering
\begin{subfigure}{.5\textwidth}
  \centering
  \includegraphics[width=\linewidth]{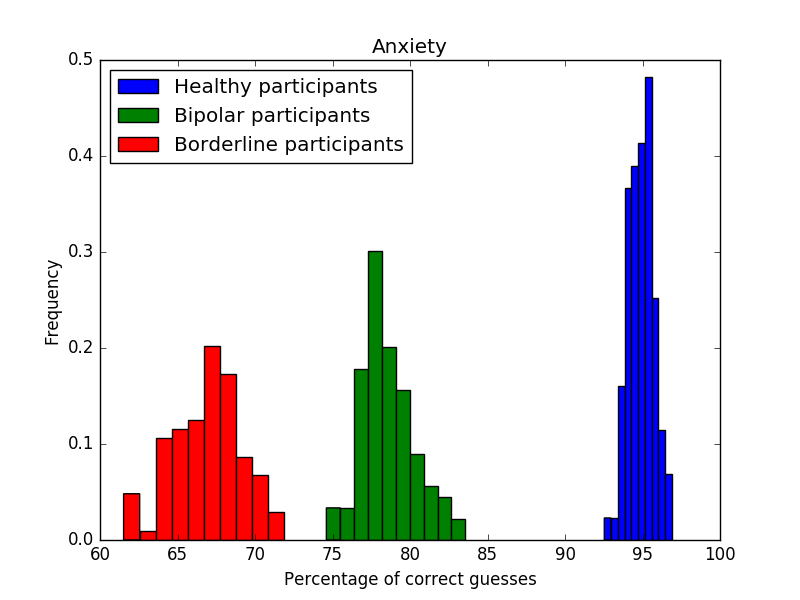}
  \caption{Anxiety.}
\end{subfigure}%
\begin{subfigure}{.5\textwidth}
  \centering
  \includegraphics[width=\linewidth]{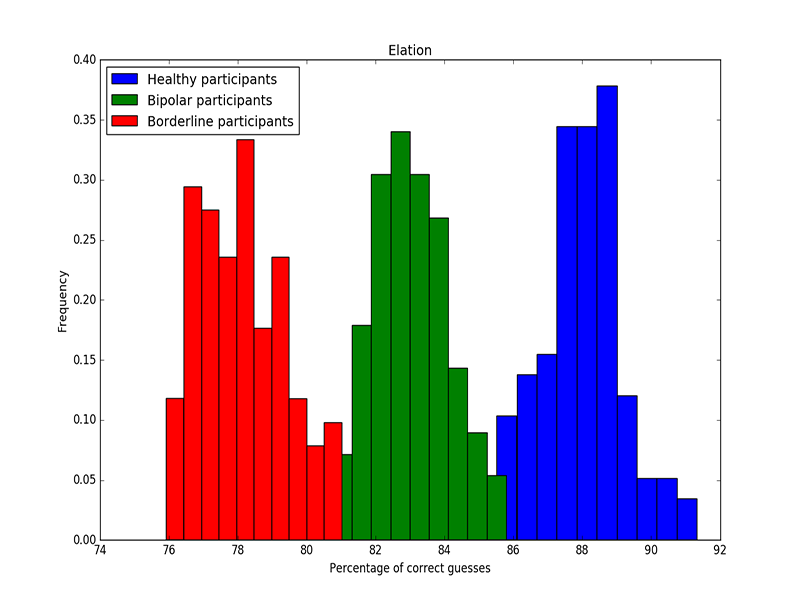}
  \caption{Elation.}
\end{subfigure}


\begin{subfigure}{.5\textwidth}
  \centering
  \includegraphics[width=\linewidth]{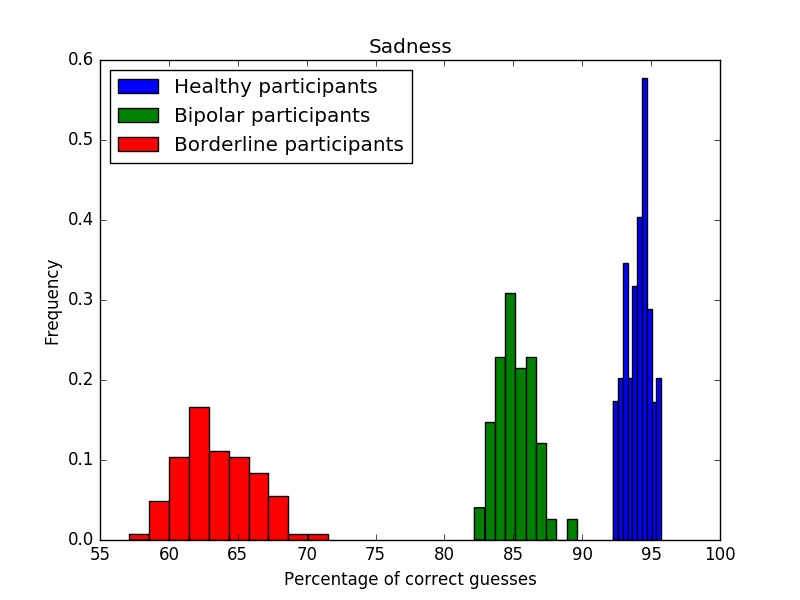}
  \caption{Sadness.}
\end{subfigure}%
\begin{subfigure}{.5\textwidth}
  \centering
  \includegraphics[width=\linewidth]{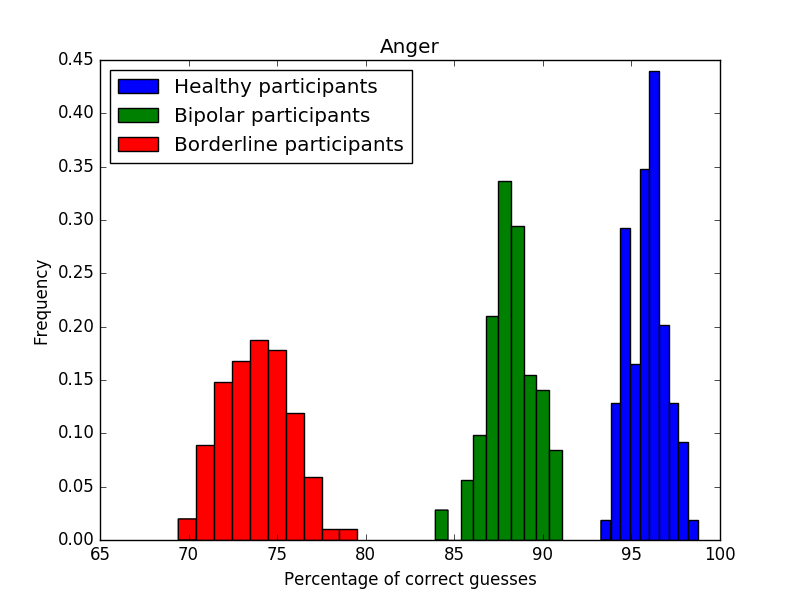}
  \caption{Anger.}
\end{subfigure}


\begin{subfigure}{.5\textwidth}
  \centering
  \includegraphics[width=\linewidth]{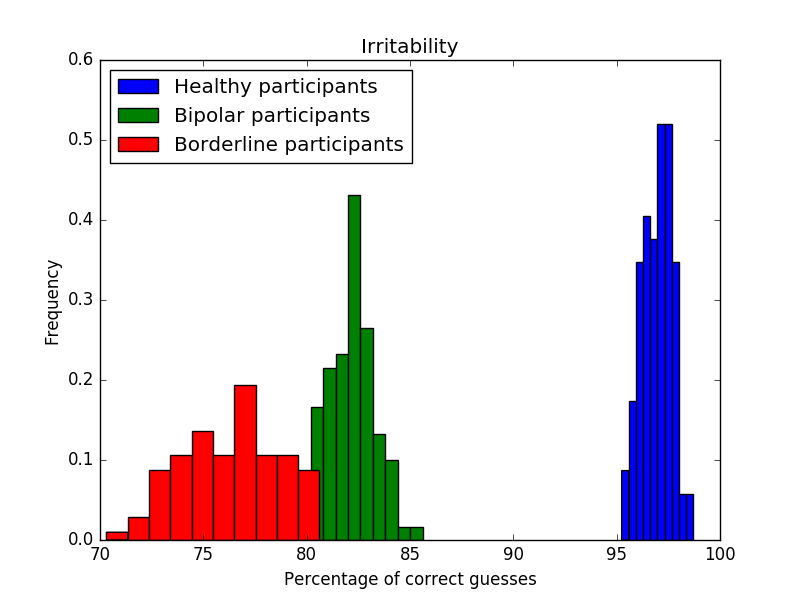}
  \caption{Irritability.}
\end{subfigure}%
\begin{subfigure}{.5\textwidth}
  \centering
  \includegraphics[width=\linewidth]{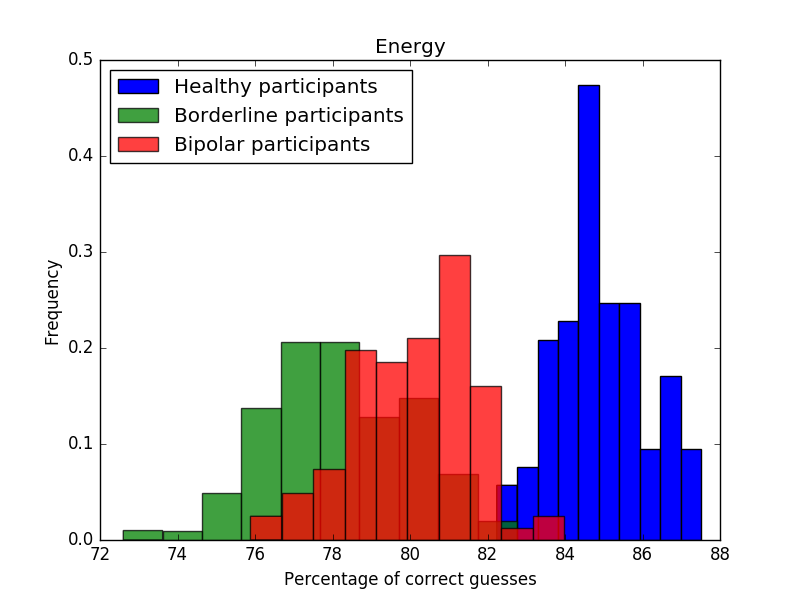}
  \caption{Energy.}
 \label{fig:energy}
\end{subfigure}

\caption{Bootstrapping of the predictive model. Notice that the three histograms are disjoint in all mood categories (except in energy scores), suggesting that there are intrinsic differences across the three clinical groups.}
\label{fig:prediction bootstrapping}
\end{figure}

On bootstrapping the concentration of the histogram of each category around its mean suggests that the model is consistent and stable. Moreover, Figure \ref{fig:prediction bootstrapping} highlights the separation between the diagnostic groups, suggesting that the three clinical groups are indeed intrinsically different.

\begin{figure}[H]
\centering
\includegraphics[height=0.9\textheight]{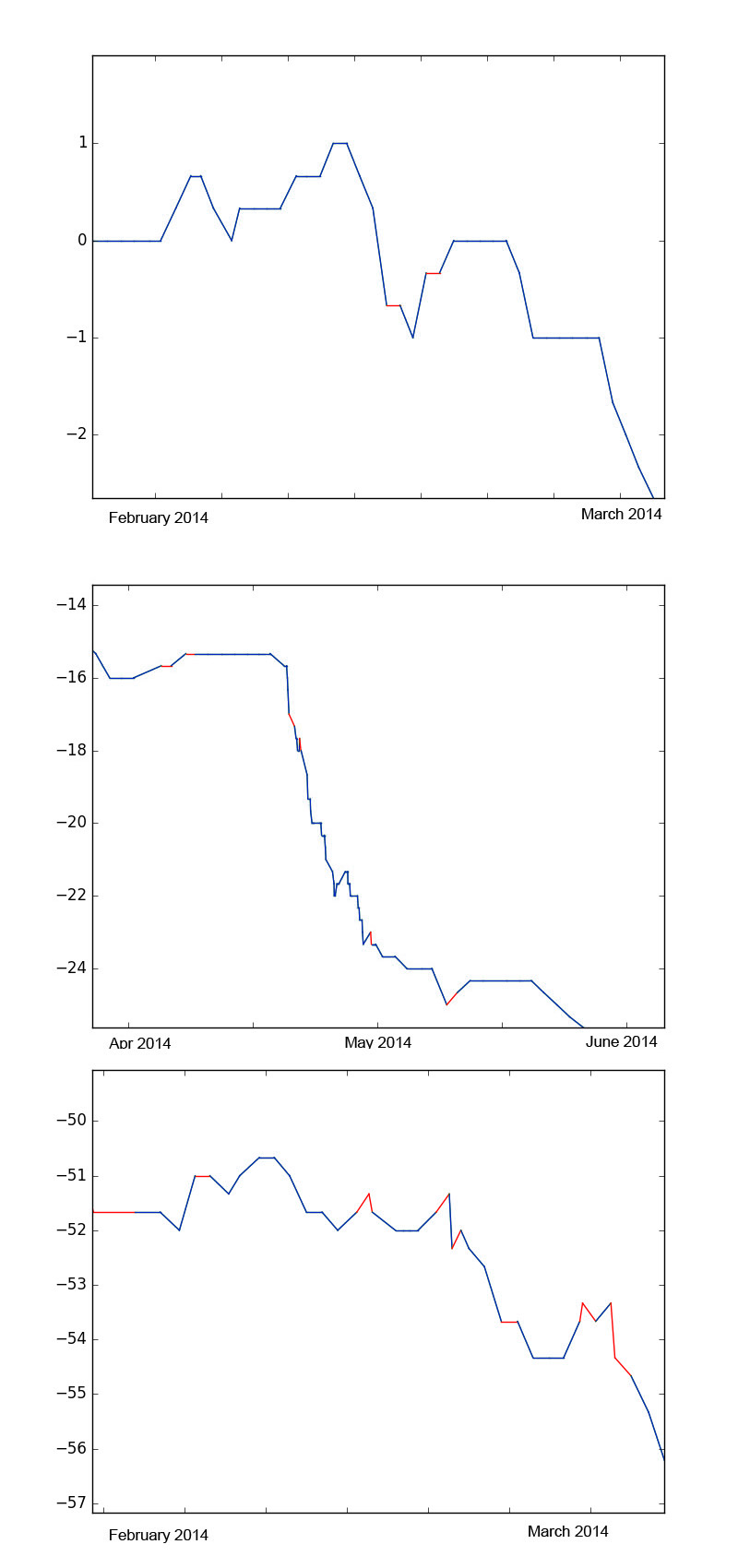}

\caption{Correct and incorrect predictions of the model. Blue indicates that the prediction is correct, and red indicates that the prediction is wrong.}
\label{fig:mood predictions}
\end{figure}

We also explored whether correct predictions of the model were derived from periods of time with a stable trend, and failures related to periods of time with rapid changes in mood. We plotted several highly oscillatory portions of the normalised path for the anxiety scores of some participants with bipolar disorder. Then, we used different colours to indicate if the predicted mood for each day was correct (see Figure \ref{fig:mood predictions}). As we see, the model correctly guesses the mood even when the path oscillates considerably.

\begin{table}[H]
\centering
\begin{tabular}{|l|cl|ll|ll|}
\hline
\multicolumn{1}{|c|}{Category} & \multicolumn{2}{c|}{Bipolar}                                                                              & \multicolumn{2}{c|}{Borderline}                                                                           & \multicolumn{2}{c|}{Healthy}                                                                             \\
\multicolumn{1}{|c|}{}         & MAE                      & \multicolumn{1}{c|}{\begin{tabular}[c]{@{}c@{}}Correct\\ predictions\end{tabular}} & \multicolumn{1}{c}{MAE} & \multicolumn{1}{c|}{\begin{tabular}[c]{@{}c@{}}Correct\\ predictions\end{tabular}} & \multicolumn{1}{c}{MAE} & \multicolumn{1}{c|}{\begin{tabular}[c]{@{}c@{}}Correct\\ predictions\end{tabular}} \\ \hline
Anxious                        & 0.96                     & 82\%                                                                           & 1.17                    & 73\%                                                                           & 0.4                     & 98\%                                                                           \\
Elated                         & 0.75                     & 86\%                                                                           & 1.03                    & 78\%                                                                           & 0.57                    & 89\%                                                                           \\
Sad                            & 0.77                     & 84\%                                                                           & 1.16                    & 70\%                                                                           & 0.41                    & 93\%                                                                           \\
Angry                          & \multicolumn{1}{c}{0.60} & 90\%                                                                           & 1.12                    & 70\%                                                                           & 0.30                    & 98\%                                                                           \\
Irritable                      & \multicolumn{1}{c}{0.84} & 84\%                                                                           & 1.15                    & 70\%                                                                           & 0.39                    & 97\%                                                                           \\
Energetic                      & \multicolumn{1}{c}{0.90} & 82\%                                                                           & 1.00                    & 75\%                                                                           & 0.69                    & 89\%                                                                           \\ \hline
\end{tabular}
\caption{Summary of the future mood prediction accuracy.}
\label{table:predict mood}
\end{table}

\section{Conclusion}

In this paper we present a novel approach to the study of psychiatric data. The signature-based model proved to be effective when it comes to making predictions about streams of 20 observations of the mood of a participant in the study. Moreover, the fact that the model was built without being polluted by the nature or structure of the data suggests that the model is robust and applicable beyond just the cohort that took part in the study.

\section{Acknowledgements}

KEAS was funded by a Wellcome Trust Strategic Award (CONBRIO: Collaborative Oxford Network for Bipolar Research to Improve Outcomes, Reference number 102616/Z). Terry Lyons is supported by the Alan Turin Institute under the EPSRC grant EP/N510129/1, and Imanol Perez is supported by a scholarship from \textit{La Caixa}.

\afterpage{
\clearpage
\begin{figure}[H]
\centering
\includegraphics[width=0.6\linewidth]{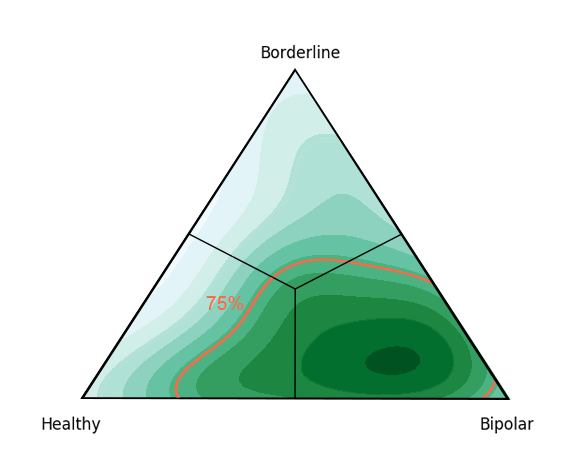}
\caption{Bipolar participants.}
\label{fig:triangle1}
\centering
\includegraphics[width=0.6\linewidth]{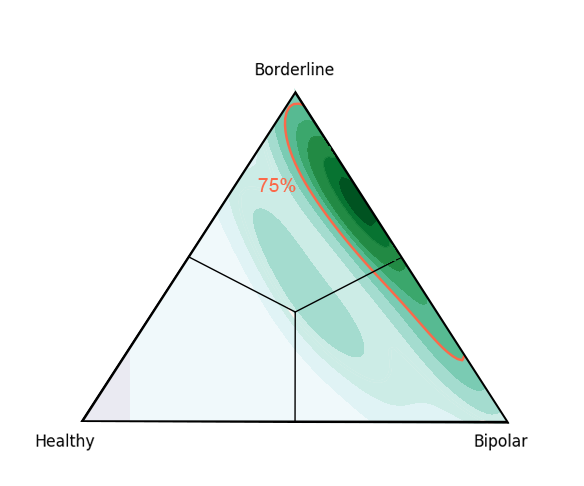}
\caption{Borderline participants.}
\label{fig:triangle2}
\centering
\includegraphics[width=0.6\linewidth]{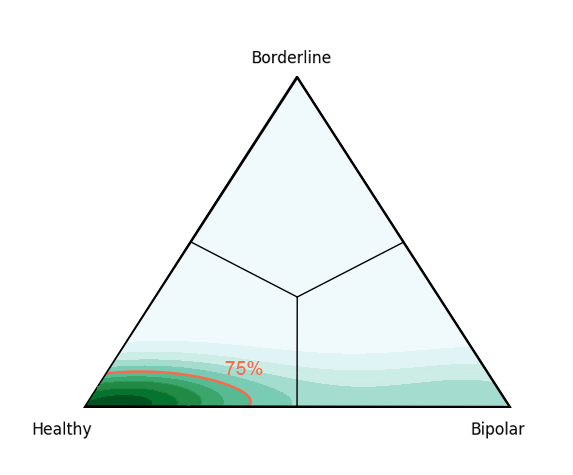}
\caption{Healthy participants.}
\label{fig:triangle3}
\end{figure}
\clearpage}

\bibliography{references}
\bibliographystyle{alpha} 

\end{document}